\documentclass[lettersize,journal]{IEEEtran}

\usepackage{amsmath,amsfonts}
\usepackage{algorithmic}
\usepackage{algorithm}
\usepackage{array}
\usepackage[caption=false,font=normalsize,labelfont=sf,textfont=sf]{subfig}
\usepackage{textcomp}
\usepackage{stfloats}
\usepackage{url}
\usepackage{verbatim}
\usepackage{graphicx}
\usepackage{cite}
\usepackage{multirow}
\begin{document}

\title{ITRE: Low-light Image Enhancement Based on  Illumination Transmission Ratio Estimation}

\author{Yu Wang, Yihong Wang, Tong Liu, Xiubao Sui, Qian Chen
	\thanks{}
	\thanks{Yu Wang(e-mail: yurowang@163.com; ).}}

\markboth{}%
{Shell \MakeLowercase{\textit{et al.}}: A Sample Article Using IEEEtran.cls for IEEE Journals}


\maketitle

\begin{abstract}
Noise, artifacts, and over-exposure are significant challenges in the field of low-light image enhancement. Existing methods often struggle to address these issues simultaneously. In this paper, we propose a novel Retinex-based method, called ITRE, which suppresses  noise and artifacts  from the origin of the model, prevents over-exposure throughout the enhancement process. Specifically, we assume that there must exist a pixel which is least disturbed by low light within pixels of same color. First, clustering the pixels on the RGB color space to find the Illumination Transmission Ratio (ITR) matrix of the whole image, which determines that noise is not over-amplified easily. Next, we consider ITR of the image as the initial illumination transmission map to construct a base model for refined transmission map, which prevents artifacts. Additionally, we design an over-exposure module that captures the fundamental characteristics of pixel over-exposure and seamlessly integrate it into the base model. Finally, there is a possibility of weak enhancement when inter-class distance of pixels with same color is too small. To counteract this, we design a Robust-Guard module that safeguards the robustness of the image enhancement process. Extensive experiments demonstrate the effectiveness of our approach in suppressing noise, preventing artifacts, and controlling over-exposure level simultaneously.  Our method performs superiority in qualitative and quantitative performance evaluations by comparing with state-of-the-art methods.
\end{abstract}


\section{Introduction}
The enhancement of low-light images can be achieved through hardware devices or software processing techniques. While hardware-based enhancement methods need to consider various factors such as noise, human vision, and economic costs, software-based approaches offer simplicity and flexibility, which have led many researchers to explore this route \cite{ren2020lr3m,li2018structure}. Low-light images typically exhibit characteristics such as low illumination, low contrast, and high levels of noise. Numerous methods have been developed to improve the quality of low-light images, broadly categorized into histogram-based \cite{coltuc2006exact,abdullah2007dynamic,pizer1990contrast,kim1997contrast,liu2019adaptive}, Retinex-based \cite{ren2020lr3m,li2018structure,jobson1997properties,jobson1997multiscale,herscovitz2004modified,xiao2013adaptive,wang2013naturalness,fu2016weighted,guo2016lime,fu2016fusion,hao2020low,ren2018lecarm}, and learning-based techniques \cite{wang2023pmsnet,li2021learning,wu2023cycle,ma2022toward, jiang2021enlightengan, wei2018deep, liu2021retinex}. Histogram-based methods have evolved from non-blocked histograms \cite{abdullah2007dynamic} to blocked-based \cite{pizer1990contrast}, non-segmented histograms to segmented-based histograms \cite{kim1997contrast}, and without gamma transformation to gamma-based histograms \cite{liu2019adaptive}. These methods largely improve images quality. However, they are still prone to over-exposure or under-exposure. In addition, the histogram-based methods are difficult to separate noise from details \cite{fu2016fusion,xu2013generalized}, leading to noise amplification.

Retinex-based models have gained considerable attention in recent years. The physical principle is to consider image as the product of illumination and reflectance, which obtains reflectance by estimating illumination, then outputs the reflectance as corrected image. Early Retinex-based methods, such as  \cite{jobson1997properties,jobson1997multiscale}, utiliz reflectance as the enhanced image. More recent variations of Retinex-based models, including those by \cite{xu2012structure,wang2013naturalness,fu2016weighted,fu2016fusion}, combine reflectance and illumination to achieve more natural results. However, these methods often incorporate gamma correction or image decomposition, introducing artifacts. The LIME method \cite{guo2016lime} extracts an initial estimation with structure-awareness from low-light images, avoiding artifacts. However, LIME fails to fundamentally address the relationship between noise and low-light images, leading to noise amplification and over-exposure. 

Artificial intelligence methods have been widely studied in recent years. Supervised-based methods tend to yield better results, but are prone to overfitting. In fact, paired datasets are not easily obtained and the robustness of such methods when applied on different data is challenging due to the limited data type in training stage. The unsupervised approaches extend robustness of deep learning models. Zero-DEC++ \cite{li2021learning} novelly defines the problem as curve estimation, which is effective in improving brightness. But it is vulnerable to be over-exposure and under-saturation when applied with different datasets. SCI \cite{ma2022toward} performs better in terms of noise suppression, but appears more obvious over-exposure. In general, they both have some limitations.

Based on the analysis above, the key challenges in low-light image enhancement lie in the suppression of noise, over-exposure, and artifacts. Existing Retinex-based methods primarily focus on noise suppression using sparsity constraints of the reflectance map, neglecting the characteristics of noise in low-light images. Furthermore, these methods tend to suppress over-exposure by relying on regularization model decomposition or synthesis of reflectance map and illumination map.
In addition, model decomposition can introduce artifacts easily. Therefore, there are some limitations.

In this paper, we follow the conventional approach to design a low-light enhancement model. Firstly, we assume the existence of a pixel with maximum illumination for each color in the input image, representing the least affected pixel by low illumination. Based on pixel color classification, we derive the initial Illumination Transmission Ratio (ITR) matrix. We then present a regularization model with over-exposure suppression function to obtain a refined illumination transmission map by using initial ITR. Additionally, we introduce a Robust-Guard (RG) module to address cases when the inter-class distance between pixels of the same color is too close, resulting in insufficient enhancement. Finally, we conduct ablation experiments and comparisons to demonstrate the effectiveness of our approach.

\begin{figure*}[!t]
	\centering
	\includegraphics[width=\linewidth]{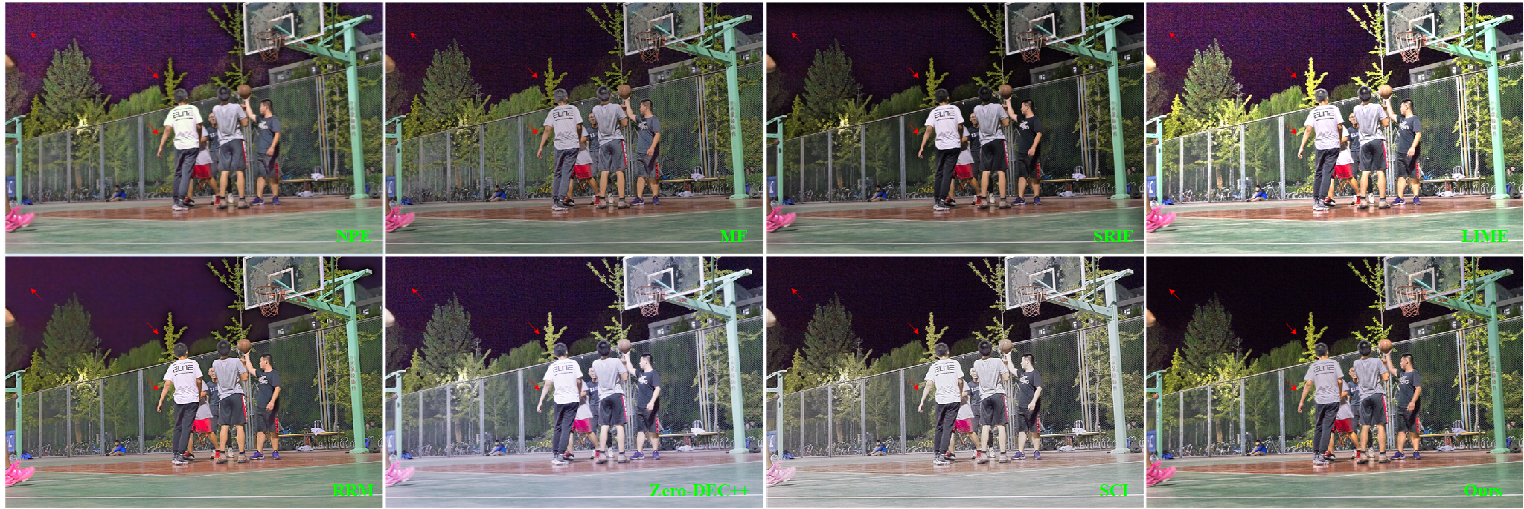}
	\caption{Comparison between our method and state-of-the-art methods. The marked regions clearly demonstrate the limitations of the compared methods, including issues such as over-exposure, noise amplification, artifacts, and low saturation, which significantly degrade the visual quality. In contrast, our method produces visually appealing results with correct exposure, absence of artifacts, minimal noise (particularly in flat, dark skies), and vibrant color representation.}
	\label{Fig.1}
\end{figure*}

In general, the contributions of this paper can be summarized as follows:
\begin{itemize}
\item We propose a novel noise suppression method based on color clustering to derive the illumination transmission ratio matrix of the ideal image, which is not found in the existing work on low-light enhancement field. Our regularization model achieves noise suppression without integrating specific noise suppression measures, as illustrated in Fig. 1.

\item We design the over-exposure suppression module from the most essential cause of the over-exposure phenomenon, and flexibly controlled the exposure level.  This design is not found in previous low-light enhancement methods.

\item We introduce a RG module to address the challenge of weak enhancement caused by a small inter-class distance between pixels of the same color. 

\item Extensive experiments are conducted, evaluating both traditional and learning-based methods. The results demonstrate the superiority of our approach in suppressing noise, over-exposure, and artifacts.
		
\end{itemize}

The remainder of the paper is organized as follows.
In Section II, the proposed ITR and base model are demonstrated in detail. In Section III, the over-exposure suppression method is described. Section IV describes overall model solver. In Section V, we introduce a RG module to guard robustness. Section VI presents extensive experiments and evaluations of different methods. Section VII demonstrates conclusion and prospect of our work.

\section{Method}

Color images convey distinct information based on the pixel color composition, i.e., an RGB image representing a fusion of various colors.  The number of colors in an image is several orders of magnitude smaller than the number of pixels \cite{NonLocalImageDehazing}. In the field of low-light image enhancement, the Retinex model serves as basic theory. In this work, we extend upon the normalized Retinex model, as described below:
\begin{equation}
\mathbf{S}=\mathbf{R}\circ \mathbf{T},  
\label{Eq.1}             
\end{equation}                                
where $\mathbf{S}$ is the degraded image (i.e., low-light image), $\mathbf{R}$ is reflectance (i.e., clear image), and T is illumination. For any pixel $\mathbf{S}_i$ in $\mathbf{S}$ , suppose its color clustering belongs to $W_j$, then there must exists a pixel in $W_j$ that is minimally affected by low light. We assume that
$\max \left\{\mathbf{T}_{w_j}\right\}=1$, then corresponding $\mathbf{R}_{w_j}=\max \left\{\mathbf{S}_{w_j}\right\}$. For all pixels belonging to the $W_j $ cluster, if we can find the illumination  ratio of these pixel points with respect to  $ \max \left\{\mathbf{T}_{w_j}\right\}$, then we can find the corresponding $\mathbf{R}$ for all pixels in $W_j$ through Eq. 1.

As discussed above, the correction of low-light images can be achieved by obtaining the relative inter-class illumination values of all pixels. We denote relative inter-class illumination as Illumination Transmission Ratio (ITR). The first key to addressing this challenge lies in color clustering. In color images, the color characteristics is determined by  r, g, b channels collectively , which are independent of its location in image. Consequently, pixel classification can be performed using 3D polar coordinates. However, for large-scale images, this classification process can be time-consuming. In a previous study on image dehazing \cite{NonLocalImageDehazing,AirlightEstimation}, the authors employed spherical coordinates to cluster pixels into different colors and utilized k-means clustering with K-D trees for efficient classification. Inspired by their approach, we have adopted this method of pixel classification in our work.

Next, we need to find the ITR matrix for all pixels with different colors. For pixels in $W_j$ , we let

\begin{equation}
	\max \left\{\mathbf{T}_{w_j}\right\}=\sqrt{\max \left\{\sum_{\mathrm{c}}\left(\mathbf{S}_{w_j}\right)^2\right.}, c \in\{r, b, g\}.
\end{equation} 

\begin{figure*}[!t]
	\centering
	\includegraphics[width=\linewidth]{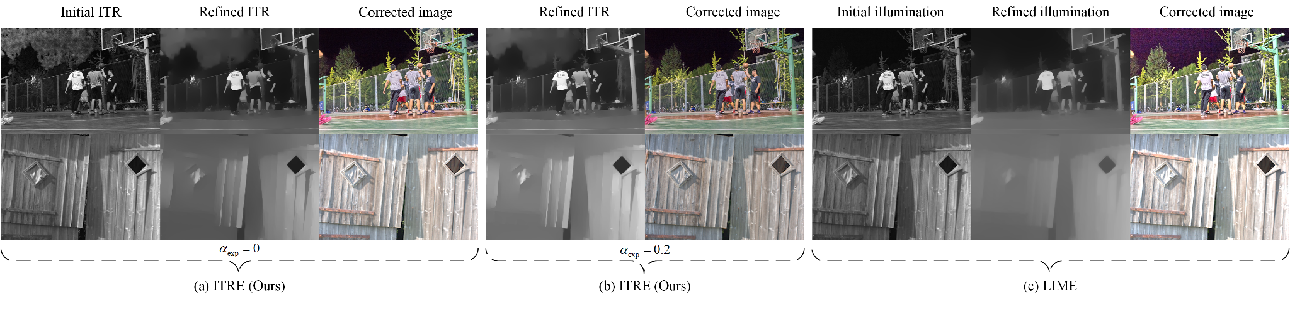}
	\caption{Comparison of intermediate results between our method and LIME.}
	\label{Fig.2}
\end{figure*}
$\max \left\{\mathbf{T}_{w_j}\right\}$ represents the maximum illumination for $W_j$ pixels in rgb color space. To prevent anomalies, we use a WLS filter \cite{farbman2008edge} for refining. Then the ITR of $W_j$ is as follows:
 
\begin{equation}
{{\overline{\mathbf{T}}}_{w_j}}={\mathbf{T}_{w_j}}/\max \left\{\mathbf{T}_{w_j}\right\}.
\label{Eq.3}
\end{equation} 
 
 The ITR of the whole image is obtained by operating Eq. \ref{Eq.3} on all pixels. There exists inaccurate case in the estimation of color. Even if the colors can be distinguished by human eye, different people are not necessarily giving the exact same classification. Therefore, we use ${{\overline{\mathbf{T}}}_w}$ as the initial illumination  transmission map and refine it by regularization.
  
    Illimination's property is broad consensus \cite{ma2022toward}. The structure of refined $\mathbf{T}$ and ${{\overline{\mathbf{T}}}_w}$ is similar, and Frobenious is often used in regularization to constrain the similarity \cite{xu2012structure}. So we complete the similarity constraint using  $\left\|\mathbf{T}-\overline{\mathbf{T}}_{{w}}\right\|_{\mathbf{F}}^2 $. A good structure of $\mathbf{T}$ is consistent with the structure of input image, i.e., the gradient of $\mathbf{T}$ varies in magnitude with the gradient of the input image \cite{wang2014variational,ng2011total}. $\ell _ {1} $  norm is piecewise smoothness, we use $\lambda\|\nabla \mathbf{T}\|_1$  to constraint structure and smoothness of $\mathbf{T}$. $\lambda$ is designed as:
  \begin{equation}
  \lambda=e^{-{\lambda_{sat}}^2}*\frac{\lambda_g}{\lambda_{gra}+eps}, 
  \label{Eq.4}
  \end{equation}
 where $\lambda_{gra}$ corresponds to the gradient of the degraded image in YUV color space. $\lambda_{sat}$ is the saturation of S. Higher saturation levels are associated with more vibrant colors, where typically require richer detail. Thus the saturation is designed to be negative here. Similarly, when dealing with larger gradients, the refined illumination component should also exhibit substantial gradients. To achieve this, we design here as the reciprocal of the gradient. $\lambda$ is jointly controlled by $\lambda_{gra}$ and$\lambda_{sat}$. $\lambda_g$ is tunable scalar and is used to adjust $\lambda$. $eps$ and $\lambda_g$ are all set  to to 0.001. Based on above, the objective function is designed as follows:

\begin{equation}
	\underset{\mathbf{T}}{\arg \min }\left\|\mathbf{T}-\overline{\mathbf{T}}_{{w}}\right\|_{\mathbf{F}}^2 +\lambda\|\nabla \mathbf{T}\|_1
	\label{Eq.5}.
\end{equation}
   
  \begin{figure}[!t]
  	\centering
  	\includegraphics[width=\linewidth]{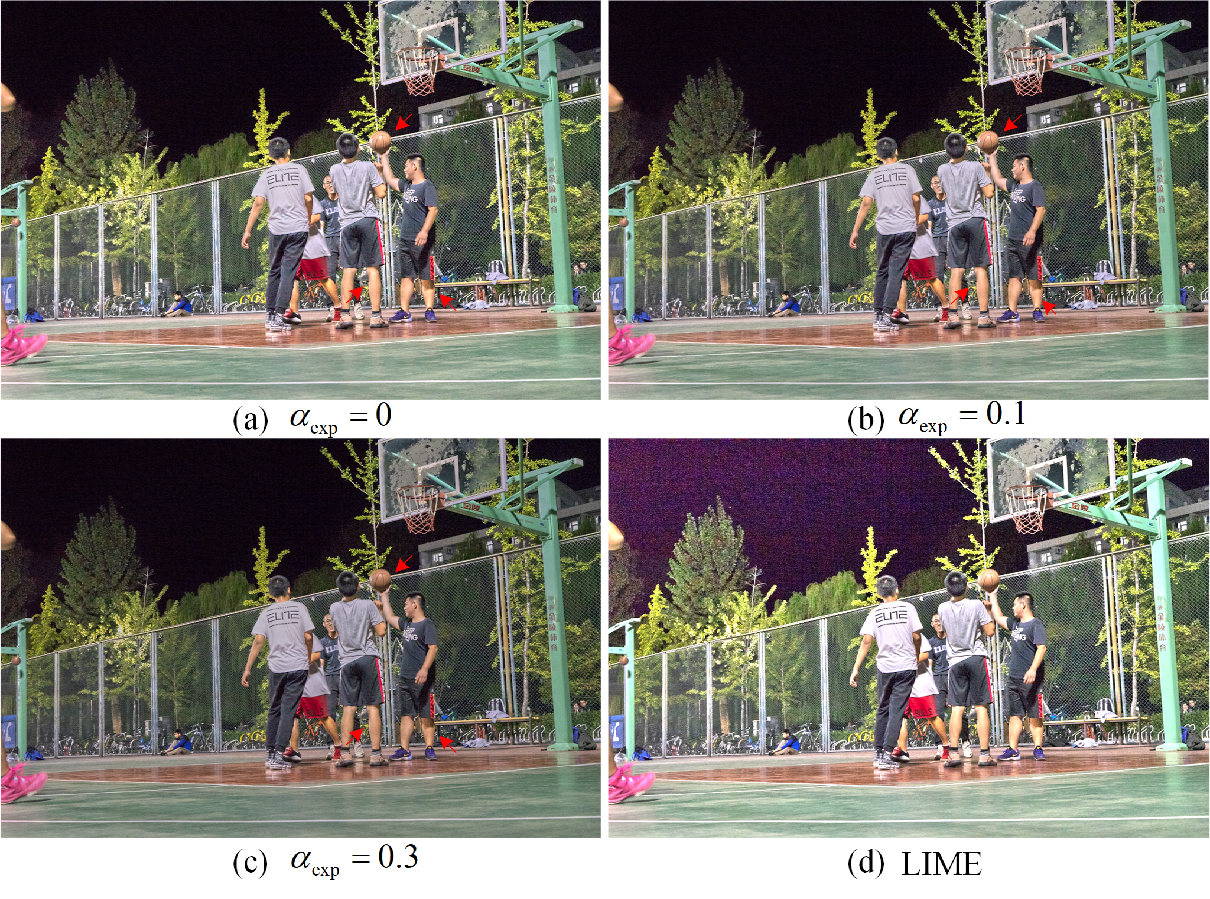}
  	\caption{Visual results of our model on different over-exposure control factor (a, b, c) and visual result of the LIME method based on illumination map estimation (d).}
  	\label{Fig.3}
  \end{figure}
\section{Over-exposure suppression}
Since standard value of refined $\mathbf{T}$ is between 0 and 1. In order to prevent anomalous values of $\mathbf{S}/\mathbf{T}$, a very small value is often added to denominator. Here we set $eps1$ to 0.1. Nevertheless, there still exists a possibility that $\mathbf{S}/(\mathbf{T}+eps1)$ is greater than 1, which increases the risk of over-exposure of the corrected image. This is not user friendly for the low brightness image enhancement applications. We design an exposure adjustment module, complementing model (\ref{Eq.5}) as follows:
\begin{equation}
	\resizebox{.93\hsize}{!}{$
	\underset{\mathbf{T}}{\arg \min }\left\|\mathbf{T}-\overline{\mathbf{T}}_{{w}}\right\|_{\mathrm{F}}^2 \quad+\lambda\|\nabla \mathbf{T}\|_1+\alpha_{\exp }\left\|{nor}\left(\frac{\mathbf{S}_{ {gray }}}{\mathbf{T}+e p s 1}-1\right)\right\|_{\mathrm{F}}^2$},
	\label{Eq.6}
\end{equation}
where $\alpha_{exp}$ is over-exposure adjustment factor. Since the estimated $\mathbf{T}$ is single channel, gray image $\mathbf{S}_{gray}$ is more suitable to represent  degraded image here. $\frac{\mathbf{S}_{gray}}{\mathbf{T}+eps1}$ refers to the reference image of reflectance. If $\frac{\mathbf{S}_{gray}}{\mathbf{T}+eps1}-1$ is larger, it means that the pixels value of reflectance is more likely to be greater than 1, i.e., the area that is prone to over-exposure. If $\frac{\mathbf{S}_{gray}}{\mathbf{T}+eps1}-1$ is smaller, it means that reflectance pixels’ value tend to be in the direction of less than 1. To simplify data processing, normalization is performed by $nor$.

The larger $\alpha_{exp}$ is, the smaller $nor(\frac{\mathbf{S}_{gray}}{\mathbf{T}+esp1}-1)$ is, so reflectance moves towards $\downarrow$1. The smaller $\alpha_{exp}$ is, the larger $nor(\frac{\mathbf{S}_{gray}}{\mathbf{T}+esp1}-1)$ is, then reflectance moves towards $\uparrow$1. As a result, $\alpha_{exp}$ achieves suppression of over-exposure phenomenon and produces better visual perception. As shown in Fig. \ref{Fig.3}, the brightness of corrected image is brighter when $\alpha_{exp}=0$. As $\alpha_{exp}$ increases, the brightness of corrected image becomes weaker. For areas with over-exposure, they become significantly darker after $\alpha_{exp}$ becomes larger (see red arrow). LIME (Fig. \ref{Fig.3} (d)) does not consider the characteristics of the low bright pixels and the nature of the exposure phenomenon, which leads to over-boosting of noise at the sky and over-exposure at several areas. Here, we do detailed experiments to illustrate the difference between LIME (based on luminance map estimation) and our model (basd on ITR estimation), shown in Fig. \ref{Fig.2}.

\begin{figure}[!t]
	\centering
	\includegraphics[width=\linewidth]{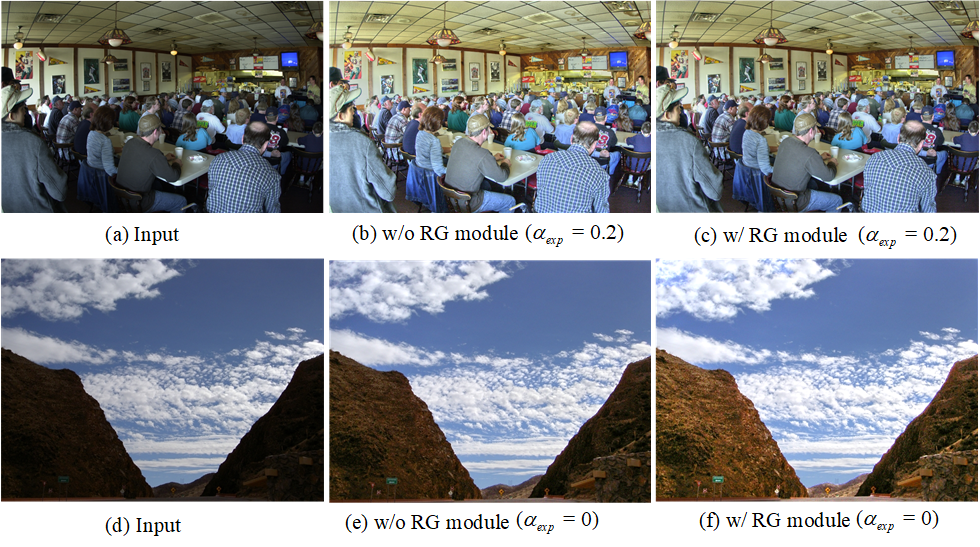}
	\caption{Comparing lightness variations on whether using RG module. }
	\label{Fig.4}
\end{figure}

\section{Model Solver}

Model (\ref{Eq.6}) contains $\ell _ {1} $ norm, and such variation functions tend to obtain optimal solutions by Alternating Direction Method of Multipliers (ADMM) \cite{guo2016lime,NonLocalImageDehazing,xu2012structure,liang2018hybrid,lin2011linearized,hong2017linear} . The solution introduces auxiliary variables to decompose the original problem into multiple subproblems. Then it is iterated through frequency domain calculations and parameter updating up to specified maximum number of iterations. By introducing auxiliary variables, Formula (6) can be rewritten as: 

\begin{equation}
		\begin{aligned}
	&\underset{\mathbf{T}}{\arg \min }\left\|\mathbf{T}-\overline{\mathbf{T}}_{w}\right\|_{\mathrm{F}}^2 +\lambda\|\mathbf{Q}\|_1
	+(\mathbf{Q}-\nabla \mathbf{T})^{\mathrm{T}} \mathbf{Y}\\
	&+\frac{\rho}{2}\|\mathbf{Q}-\nabla \mathbf{T}\|_{\mathrm{F}}^2+\alpha_{\exp }\left\|{nor}\left(\frac{\mathbf{S}_{ {gray }}}{\mathbf{T}+e p s 1}-1\right)\right\|_F^2,\\
		&\text { s.t. } \quad \mathbf{Q}=\nabla\mathbf{T}.
		\end{aligned}
	\label{Eq.7}
\end{equation}

Where $\nabla$ includes horizontal gradient and vertical gradient operator. $\mathbf{Y}$ is Lagrangian dual variable. The solution of (\ref{Eq.7}) is decomposed into two subproblems.

$\mathbf{T}$ \textbf{subproblem}: By omitting terms not related $\mathbf{T}$, we get 

\begin{equation}
	\begin{aligned}
		&\underset{\mathbf{T}}{\arg \min }\left\|\mathbf{T}-\overline{\mathbf{T}}_{w}\right\|_{\mathrm{F}}^2 \quad
		+(\mathbf{Q}-\nabla \mathbf{T})^{\mathrm{T}} \mathbf{Y}\\
		&+\frac{\rho}{2}\|\mathbf{Q}-\nabla \mathbf{T}\|_{\mathrm{F}}^2+\alpha_{\exp }\left\|{nor}\left(\frac{\mathbf{S}_{ {gray }}}{\mathbf{T}+e p s 1}-1\right)\right\|_F^2.
	\end{aligned}
\end{equation}

We notice that (8) must includes many mathematical operations of matrices inversion, which need high computational cost. The process can be solved by 2D fast Fourier transformation (FFT). Hence, $\mathbf{T}^{(t+1)}$ is iteratively updated by

\begin{equation}
\mathbf{T}^{(t+1)}\gets\mathcal{F}^{-1}\left(\frac{\mathcal{F}\left({\overline{\mathbf{T}}}_{w}+\rho^{(t)}\mathbf{D}^T\left(\mathbf{Q}^{(t)}+\frac{\mathbf{Y}^{(t)}}{\rho^{(t)}}\right)\right)+\mathcal{F}(\mathcal{\mathbf{E}})}{1+{\rho^{(t)}\sum_{d\in{h,v}}\mathcal{F}}^*\left(\mathbf{D}_d\right)\circ\mathcal{F}\left(\mathbf{D}_d\right)}\right).
\label{Eq.9}
\end{equation}

\begin{figure}[!ht]
	\centering
	\includegraphics[width=\linewidth]{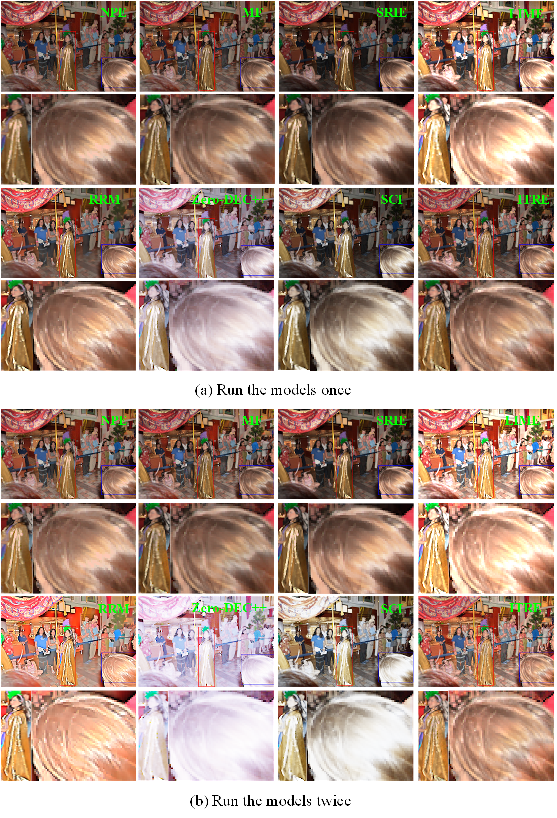}
	\caption{Comparison of stability based MIT-Adobe 5K datasets \cite{bychkovsky2011learning}. }
	\label{Fig.5}
\end{figure}
Where $\mathbf{D}$ is discrete gradient operator in vertical and horizontal directions. $\mathcal{F}^{-1}$, $\mathcal{F}$ and $\mathcal{F}^*$ denote inverse FFT, Fourier Transform, conjugate Fourier Transform, respectively.  $\mathcal{F}(\mathbf{E})$ corresponds to the exposure control items in (\ref{Eq.6}).  At the first iteration, $\mathbf{E}$ is initialized to 0. When the number of iterations is bigger than 1, the over-exposure module detects the exposure level and participates in the generation of $\mathbf{T}$, and it will be described in detail latter.

$\mathbf{Q}$ \textbf{subproblem}: Dropping terms not related $\mathbf{Q}$, we get 

\begin{equation}
	\mathbf{Q}^{(t+1)} \leftarrow \underset{\mathbf{Q}}{\arg \min }~ \lambda\|\mathbf{Q}\|_1+(\mathbf{Q}-\nabla \mathbf{T})^{\mathrm{T}} \mathbf{Y}+\frac{\rho}{2}\|\mathbf{Q}-\nabla \mathbf{T}\|_{\mathbf{F}}^2,
\end{equation}
the solution can be easily obtained by soft-shrinkage operation like:
\begin{equation}
	\mathbf{Q}^{({t}+1)}=\mathcal{T}_{\lambda / \rho^{({t})}}\left(\nabla \mathbf{T}^{({t}+1)}-\mathbf{Y}^{(t)} / \rho^{({t})}\right),
	\label{Eq.11}
\end{equation}
where $\mathcal{T}_{\lambda / \rho^{({t})}}$ means soft-shrinkage symbol, $\mathcal{T}_{\lambda / \rho^{({t})}}(x)=\operatorname{sign}(x) \cdot \max \left(|x|-\lambda / \rho^{({t})}, 0\right)$.
Similar to the above solution, Y and $\rho$ can be solved by (\ref{Eq.12}) , as follows:

\begin{equation}
	\begin{aligned}
		& \mathbf{Y}^{({t}+1)} \leftarrow \mathbf{Y}^{{(t)}}+\rho^{({t})}\left(\mathbf{Q}^{({t}+1)}-\nabla \mathbf{T}^{({t}+1)}\right), \\
		& \rho^{({t}+1)} \leftarrow p \cdot \rho^{({t})}, p>1,
	\end{aligned}
\label{Eq.12}
\end{equation}
finally, after completing the above solution, calculate 
$\mathbf{E}^{(t+1)}$ :

\begin{equation}
	\mathbf{E}^{({t}+1)} \leftarrow \alpha_{\exp } {nor}\left(\frac{\mathbf{S}_{ {gray }}}{\mathbf{T}^{(t+1)}+e p s 1}-1\right).
	\label{Eq.13}
\end{equation}

After refined ITR is estimated, we follow Retinex model (\ref{Eq.1}) to get enhanced image. To understand our model more intuitively, we present the overall solution in Algorithm 1.

\begin{algorithm}[!ht]
	\caption{Pseudo-Code for Model Framework}
	\begin{algorithmic}
		\STATE 
		\STATE {\textbf{ITR Estimation:}}
		\STATE 
		Cluster pixels’ colors in spherical coordinates, find the initial ITR   by Eq. \ref{Eq.3}.
		\STATE {\textbf{Input}} $\mathbf{S}$, $\mathbf{S}_{gray}$,\ $\alpha_{exp}$, $\lambda$ (Eq. \ref{Eq.4}).
		\STATE {\textbf{Initialization}} $\mathbf{E}^{(0)}$=$\mathbf{Q}^{(0)}$=$\mathbf{Y}^{(0)}$=0,
		 t=0, 
		 while not converged do
		 \STATE \hspace{0.5cm} Update $\mathbf{T}^{(t+1)}$ by Eq. \ref{Eq.9};
		\STATE \hspace{0.5cm}  Update  $\mathbf{Q}^{(t+1)}$ by Eq. \ref{Eq.11};
		\STATE \hspace{0.5cm} Update $\mathbf{Y}^{(t+1)}$ and $\rho^{({t}+1)}$ by Eq. \ref{Eq.12};
		\STATE \hspace{0.5cm} Update $\mathbf{E}^{(t+1)}$ by Eq. \ref{Eq.13};
		\STATE \hspace{0.5cm} $t$=$t$+1;
		\STATE \hspace{0.5cm}end
		\STATE {\textbf{ Low-light Correction:}}
		\STATE Calculate $\mathbf{R}$  through Eq. \ref{Eq.1}.

	\end{algorithmic}
	\label{alg1}
\end{algorithm}

\begin{figure*}[!ht]
	\centering
	\includegraphics[width=0.8\linewidth]{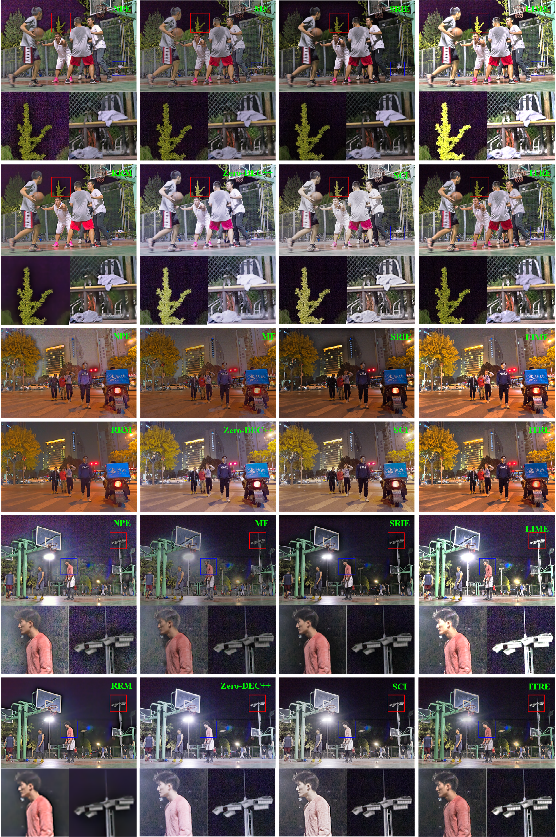}
	\caption{Visual comparison between ours and competitors on typical low-light dataset. All the results are without RG module. }
	\label{Fig.6}
\end{figure*}

\begin{figure*}[!ht]
	\centering
	\includegraphics[width=0.8\linewidth]{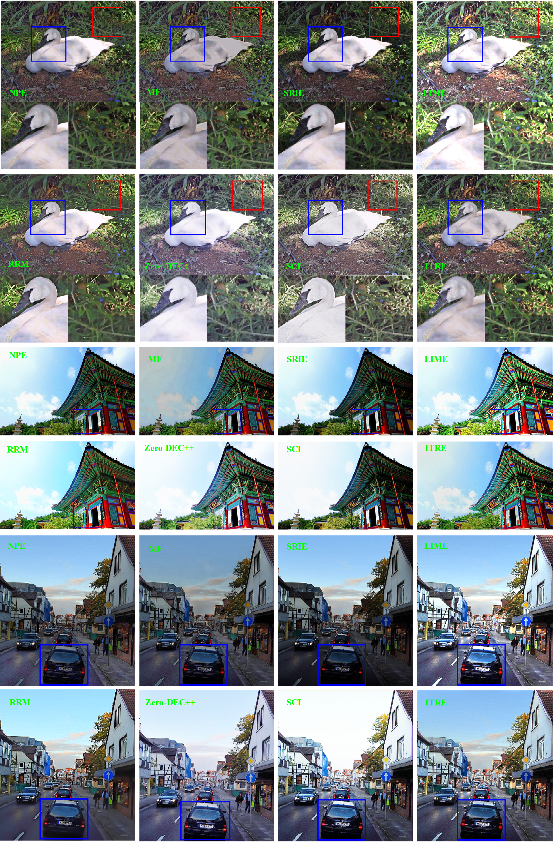}
	\caption{Visual comparison between ours and competitors on typical low-light dataset. All the results are without RG module. }
	\label{Fig.7}
\end{figure*}
\section{Robust-Guard module Setting}

For input image, ${\overline{\mathbf{T}}}_{w_j}$ tends to 1 if the pixels in the $W_j$ cluster are too close, this case leads to weak illumination  boost, as shown in Fig. \ref{Fig.4}(e). Therefore, the adjustment of ${\overline{\mathbf{T}}}_w$ is needed. We design a Robust-Guard (RG) module after ITR estimation. The v-channel of HSV color space has light invariance, so our design in the v-channel. Neighborhood maximum extraction is performed on the v-channel image $\mathbf{S}_{hsv-v}$, as follows:

\begin{equation}
	{\overline{\mathbf{T}}}_{tmp}(x,y)=\max \left\{\mathbf{S}_{hsv-v}(p,q)\right\},
\end{equation}
where ${(p,q)}$ is the neighborhood of the pixel ${(x,y)}$. The pixels in neighborhood can be roughly considered as same color, so ${\overline{\mathbf{T}}}_{tmp}$ can be regarded as rough illumination  map. Similarly, here use a WLS filter for refining. Since ${\overline{\mathbf{T}}}_{tmp}$ is extracted directly from original image, so the hue of it is darker than ${\overline{\mathbf{T}}}_w$. We use histogram matching \cite{coltuc2006exact,thomas2011histogram} by Eq. \ref{Eq.15} between  ${\overline{\mathbf{T}}}_{tmp}$ and ${\overline{\mathbf{T}}}_w$, the hue of ${\overline{\mathbf{T}}}_w$ is adjusted closer to ${\overline{\mathbf{T}}}_{tmp}$.

\begin{equation}
\overline{\mathbf{T}}_w^{\prime}=\mathbf{H}_{match}({\overline{\mathbf{T}}}_w,{\overline{\mathbf{T}}}_{tmp}). 
\label{Eq.15}
\end{equation}

The histogram matching has been integrated into existing matlab software. Then replace $\overline{\mathbf{T}}_w$ with $\overline{\mathbf{T}}_w^{\prime}$ in Eq. \ref{Eq.7}.  Fig. \ref{Fig.4}(e) and (f) show the difference between w/ and w/o RG module. However, for images where the pixels’ color are not dense enough (Fig. \ref{Fig.4}(a)), the RG module does not affect illumination  strength (Fig.\ref{Fig.4}(b) and (c)). When the image has few color varieties and pixels values with same color have small variance, there exist weak enhancement (Fig. \ref{Fig.4}(e)). But the RG module guards the enhancement (Fig. \ref{Fig.4}(f)). The experiment shows that the RG module guards illumination  enhancement.
It is worth noting that for normal low light scenes (not extreme darkness), without RG module can also ensures sufficient illumination enhancement capability.

\begin{figure}[H]
	\centering
	\includegraphics[width=\linewidth]{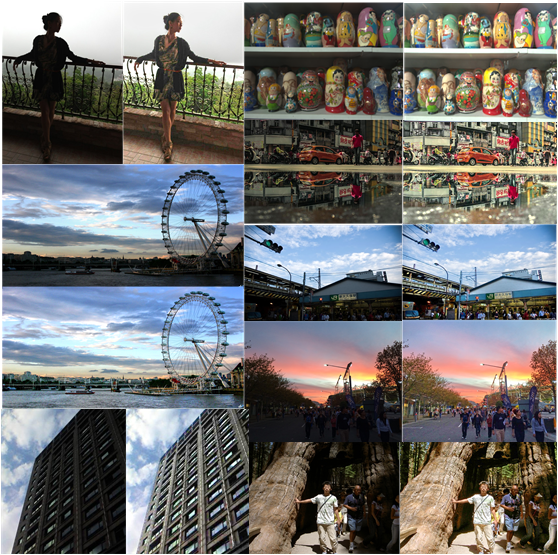}
	\caption{More results by our method. All the results are with RG module. }
	\label{Fig.8}
\end{figure}
\section{Experiment}
In this section, some important setting and results are demonstrated. We first describe the configurations of all models. Second, we compare them in terms of stability and noise. Then, we compare the subjective results and perform objective metrics tests.

\subsection{Implementation details}

We compare with 5 traditional methods and 2 recently proposed artificial intelligence methods. 5 traditional methods are NPE \cite{wang2013naturalness}, MF \cite{fu2016fusion}, SRIE \cite{fu2016weighted}, LIME \cite{guo2016lime} and RRM \cite{li2018structure} respectively. 2 artificial intelligence methods are Zero-DEC++ \cite{li2021learning} and SCI \cite{ma2022toward}. The codes of all compared algorithms are available on the web. 
All experiments are performed in MATLAB R2019a with 32 GB RAM and 12th Gen Intel(R) Core(TM) i7-12700K.

For the objective metric tests (Table \ref{Tab:1}), the parameters of proposed models and comparison models are kept constant. Too many parameters are cumbersome to adjust, which significantly influence the model are  $\alpha_{exp}$ and RG module. In our tests, we mainly adjust $\alpha_{exp}$ and RG module,  getting desired results. For the subjective test, w/o RG module, $\alpha_{exp}=0.25$ are set in Fig. \ref{Fig.5}. Fig. \ref{Fig.6} and \ref{Fig.7} are w/o RG module. In order to highlight the over-exposure suppression module more intuitively, $\alpha_{exp}$ are set to {0.1, 0, 0.3, 0.2, 0, 0} from the 1st to 6th images in  Fig. \ref{Fig.6} and \ref{Fig.7}.

\begin{table*}[]
	\centering

	\caption{Objective evaluation on low-light datasets DICM, LIME, NPE and MF.}
	\tiny
	\label{Tab:1}
	\renewcommand\arraystretch{1.5}
	
	\resizebox{\linewidth}{!}{
		\begin{tabular}{|cc|c|c|c|c|c|c|c|c|}

			\hline
			\multicolumn{2}{|l|}{\multirow{2}{*}{Method}}       & \multirow{2}{*}{NPE} & \multirow{2}{*}{MF} & \multirow{2}{*}{SRIE} & \multirow{2}{*}{LIME} & \multirow{2}{*}{RRM} & \multirow{2}{*}{Zero-DEC++} & \multirow{2}{*}{SCI} & \multirow{2}{*}{Ours} \\
			\multicolumn{2}{|l|}{}                              &                      &                     &                       &                       &                      &                             &                      &                       \\ \hline
			\multicolumn{1}{|c|}{\multirow{4}{*}{DICM}} & EME↑  & \textbf{88.9402}              & 86.2982             & 87.1266               & \textbf{90.3609}               & 27.6694              & 65.4914                     & 85.7028              & \textbf{87.6519}               \\ \cline{2-10} 
			\multicolumn{1}{|c|}{}                      & DE↑   & 6.678                & 6.6214              & 6.6828                & \textbf{6.8141}                & 6.7782               & \textbf{7.016}                       & 6.1749               & \textbf{6.8013}                \\ \cline{2-10} 
			\multicolumn{1}{|c|}{}                      & NIQE↓ & 3.4304               & 3.2324              & \textbf{3.092}                & 3.559                 & 3.3186               & \textbf{2.6509}                      & \textbf{3.1745}              & 3.3004                \\ \cline{2-10} 
			\multicolumn{1}{|c|}{}                      & PIQE↓ & 36.2486              & \textbf{31.8063}             & 34.6276               & 38.6683               & 42.5862              & \textbf{26.6375}                     & \textbf{29.307}               & 33.9231               \\ \hline
			\multicolumn{1}{|c|}{\multirow{4}{*}{LIME}} & EME↑  & 74.4371              & 75.3797             & 74.5778               & \textbf{78.0302}               & 29.5823              & 73.6806                     & \textbf{75.5075}              & \textbf{76.4803}               \\ \cline{2-10} 
			\multicolumn{1}{|c|}{}                      & DE↑   & 6.7427               & 6.6906              & 6.7031                & \textbf{7.0865}                & 6.8055               & \textbf{7.0063}                     & 6.2974               & \textbf{6.9861}                \\ \cline{2-10} 
			\multicolumn{1}{|c|}{}                      & NIQE↓ & 3.84                 & \textbf{3.5976}             & \textbf{3.463}                & 4.2076                & 4.0646               & 3.8881                      & 4.1382               & \textbf{3.8767}                \\ \cline{2-10} 
			\multicolumn{1}{|c|}{}                      & PIQE↓ & 36.4812              & \textbf{35.2572}             & \textbf{34.7532}              & 39.4609               & 44.5976              & 37.3695                     & 38.6532              & \textbf{36.5375}               \\ \hline
			\multicolumn{1}{|c|}{\multirow{4}{*}{NPE}}  & EME↑  & 47.9715              & 48.2989             & \textbf{48.4597}               & \textbf{49.0242}              & 19.285               & 31.513                      & 48.3958              & \textbf{48.6956}               \\ \cline{2-10} 
			\multicolumn{1}{|c|}{}                      & DE↑   & \textbf{7.2588}               & 7.0503              & \textbf{7.224}                & 7.1762                & 7.0895               & 7.2032                      & 6.9817               & \textbf{7.3391}                \\ \cline{2-10} 
			\multicolumn{1}{|c|}{}                      & NIQE↓ & 3.3042               & \textbf{3.2754 }             & \textbf{3.1225}               & 3.6026                & 4.6142               & \textbf{3.2919}                      & 3.3631               & 3.4121                \\ \cline{2-10} 
			\multicolumn{1}{|c|}{}                      & PIQE↓ & \textbf{33.0718}             & \textbf{31.8361}            & \textbf{34.5226}               & 37.4361               & 61.051               & 37.101                      & 35.1166              & 35.0092               \\ \hline
			\multicolumn{1}{|c|}{\multirow{4}{*}{MF}}   & EME↑  & 46.5234              & \textbf{46.7645}             & 46.4599               & \textbf{47.7578}              & 20.8585              & 32.5819                     & 45.667               & \textbf{47.0558}               \\ \cline{2-10} 
			\multicolumn{1}{|c|}{}                      & DE↑   & 6.5953               & 6.575               & 6.6033                & \textbf{6.9724}                & 6.6317               & \textbf{6.8889}                    & 5.8124               & \textbf{6.8871}                \\ \cline{2-10} 
			\multicolumn{1}{|c|}{}                      & NIQE↓ & 3.3503               &     \textbf{3.3339}              & \textbf{3.1109}               & 3.4125                & 3.7104               & 3.4412                      & 3.5558               & \textbf{3.32}                  \\ \cline{2-10} 
			\multicolumn{1}{|c|}{}                      & PIQE↓ & \textbf{41.7006}            & \textbf{40.6526}            & 43.1725               & 43.639                & 49.6911              & 46.57                       & 46.6144              & \textbf{42.059}                \\ \hline
	\end{tabular}}
\end{table*}
\subsection{Comparisons}
\textbf{Stability}. In Fig. 4, results for once-run and twice-run of all models are compared. When models are run once, LIME has high illumination  permeability but with local over-exposure. The saturation of Zero-DEC++ and SCI is low. The proposed method and the remaining comparison algorithms exist no over-exposure. When models are run twice, our model further increases the illumination  while suppressing local over-exposure. This is because the over-exposure pixels are adjusted in each iteration of Eq. \ref{Eq.6}. NPE, MF, and SRIE do not change significantly, while the rest of the models increase the over-exposure strength. The difference is that Zero-DEC++ tends to global over-exposure and the other three models tend to local over-exposure. It can be seen that our model suppresses over-exposure while enhancing illumination. This demonstrates high stability of the proposed model.

\textbf{Noise}. Low-light images are enhanced with inevitable amplification of noise. The 1st, 2nd, and 3rd images in Figure 6 are from the DARK FACE dataset taken at night \cite{yang2020advancing}. Obviously, the proposed method and SRIE have the weakest amplification of noise. However, SRIE produces significant artifacts, and our method avoids artifacts due to designing from independent pixels perspective. In general,  the areas in low-light image include more noise where they are more darker. We achieve the denoising results without using filtering and without constraining the sparsity of the reflectance map, simply by controlling the proportion of noise enhancement

\textbf{Overall comparison}. All low-light images which are tested in this paper can be found in \cite{wang2013naturalness}, \cite{guo2016lime}, \cite{fu2016fusion}, \cite{lee2012contrast}, \cite{bychkovsky2011learning}, \cite{yang2020advancing}.  Fig. \ref{Fig.6} and  \ref{Fig.7} shows the corresponding results. LIME tends to have high brightness but gets over-exposure in details, while Zero-DEC++ and SCI also suffer from over-exposure. The difference is that LIME and SCI tend to be locally over-exposed, while Zero-DEC++ tends to be more globally over-exposed. In terms of noise, noise amplification is inevitable, but the proposed method exhibits low noise amplification and the correction results have high saturation. In terms of edge artifacts, NPE, MF, SRIE, and RRM all produce artifacts, and our method does not produce artifacts.   
Thus, for subjective evaluation, our algorithm outperforms the comparison algorithms. Fig. \ref{Fig.8} provides more results for classic pictures by proposed method.

\begin{figure}[!h]
	\centering
	\includegraphics[width=\linewidth]{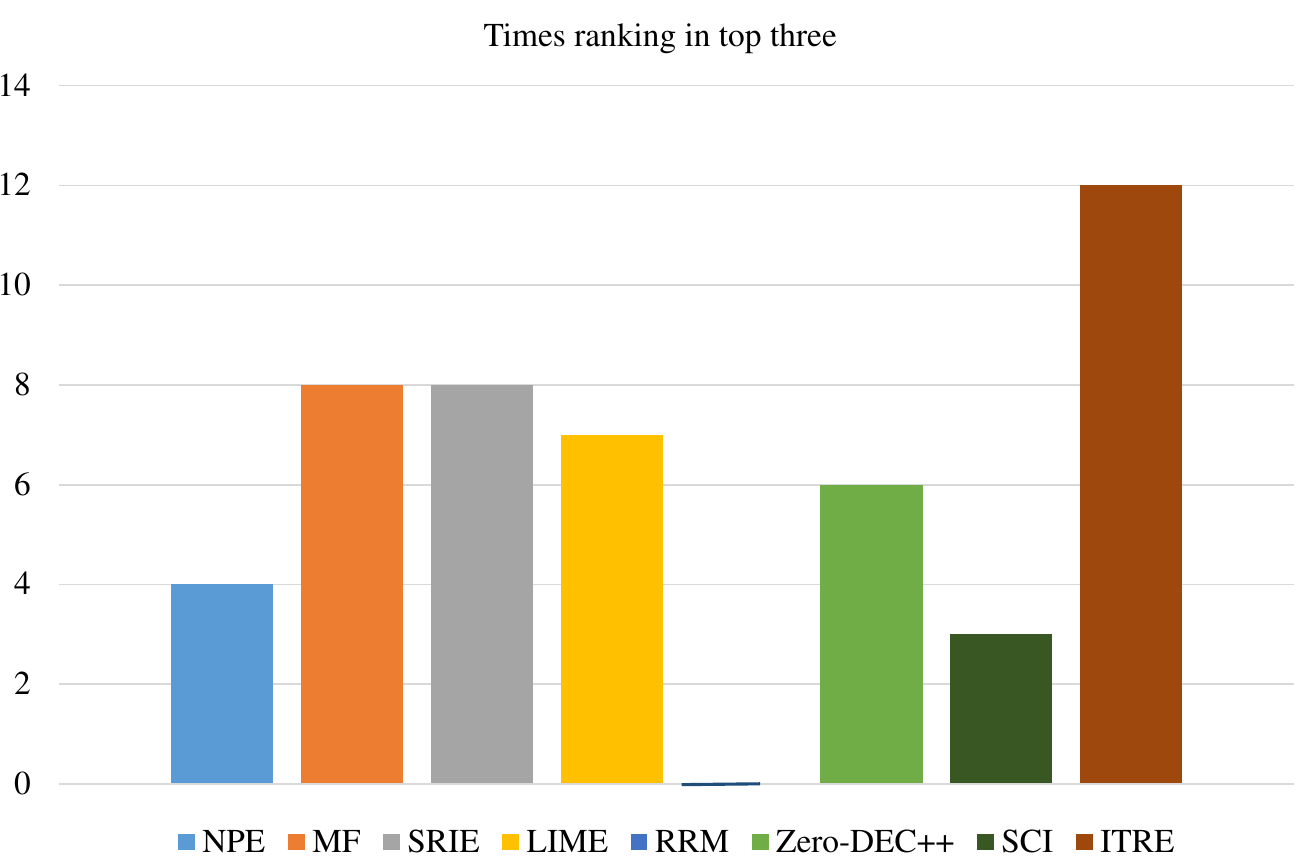}
	\caption{The times of different methods rank in  top three on EME, DE, NIQE and PIQE. }
	\label{Fig.9}
\end{figure}

We choose four non-reference metrics, EME \cite{agaian2007transform}, DE \cite{shannon1948mathematical}, NIQE \cite{mittal2012making} and PIQE \cite{venkatanath2015blind}. The larger the EME, the higher contrast the image is. The larger the DE, the more information the image is. Smaller NIQE and PIQE represent that the image has higher naturalness and it is closer to human perception, respectively. Table \ref{Tab:1} shows the metrics of all models tested in the public datasets which downloaded from websites, including 44 images in DICM \cite{lee2012contrast}, 10 images in LIME, 64 images in NPE, and 14 images in MF, for a total of 132 images. Multiple metrics are closer to the real effectiveness of models than single metric. We record the four metrics corresponding to all models in all datasets and count the number of ranking top three for all models. According to Fig. \ref{Fig.9} and Tab. \ref{Tab:1}, our model ranks first. As shown above, our model obtains more friendly visual results and better metrics compared with state-of-art methods.



\section{Conclusion}
In conclusion, existing low-light image enhancement methods often struggle to address the issues of noise, artifacts, and over-exposure simultaneously. In contrast, our proposed method effectively tackles these challenges in a unified framework. We  consider the strongest illumination value among pixels of the same color as the reference. This strategy enables us to control the enhancement ratio of similar color pixels, resulting in effective noise suppression, particularly in low illumination regions.  The initial ITR is pixel-based and with structural awareness, so it is not prone to appear artifacts when building a model based on it.  Moreover, we design an over-exposure module that effectively controls the level of over-exposure by addressing the fundamental phenomenon underlying pixel over-exposure. Additionally, our RG module ensures the robustness of image enhancement, even in scenarios where the image contains few pixel color types or the variance of pixel values with the same color is small.
Extensive comparisons with state-of-the-art conventional and artificial intelligence methods demonstrate the superiority of our approach in terms of noise suppression, artifact prevention, and over-exposure control. While our proposed method represents a significant advancement in low-light image enhancement, further research is warranted to develop more intelligent and robust algorithms that can achieve best results across diverse datasets under the same parameter configuration.

\bibliographystyle{IEEEtran}
\bibliography{IIEUIIP}

\newpage

\vfill

\end{document}